\documentclass[10pt,twocolumn,letterpaper]{article}
\usepackage{cvpr}
\usepackage{times}
\usepackage{epsfig}
\usepackage{graphicx}
\usepackage{amsmath}
\usepackage{amssymb}
\usepackage{rotating}
\usepackage{algorithm}
\usepackage{algpseudocode}
\usepackage{subcaption}
\usepackage{array}
\usepackage{multirow}

\newcommand*\mymin[1][]{\min_{#1}\,}
\newcommand*\mymax[1][]{\max_{#1}\,}

\usepackage[pagebackref=true,breaklinks=true,letterpaper=true,colorlinks,bookmarks=false]{hyperref}

\cvprfinalcopy 

\ifcvprfinal\fi
\begin{document}


\title{Generate To Adapt: Aligning Domains using Generative Adversarial Networks}

\author{Swami Sankaranarayanan \thanks{First two authors contributed equally}
\quad Yogesh Balaji $^{*}$ \quad Carlos D. Castillo \quad Rama Chellappa \\
\\
UMIACS, University of Maryland, College Park 
}

\maketitle

\begin{abstract}
Domain Adaptation is an actively researched problem in Computer Vision. In this work, we propose an approach that leverages unsupervised data to bring the source and target distributions closer in a learned joint feature space. We accomplish this by inducing a symbiotic relationship between the learned embedding and a generative adversarial network. This is in contrast to methods which use the adversarial framework for realistic data generation and retraining deep models with such data. We demonstrate the strength and generality of our approach by performing experiments on three different tasks with varying levels of difficulty: (1) Digit classification (MNIST, SVHN and USPS datasets) (2) Object recognition using OFFICE dataset and (3) Domain adaptation from synthetic to real data. Our method achieves state-of-the art performance in most experimental settings and by far the only GAN-based method that has been shown to work well across different datasets such as OFFICE and DIGITS.
\end{abstract}
\section{Introduction}
\label{sec:intro}

\begin{figure*}
\includegraphics[width=\textwidth]{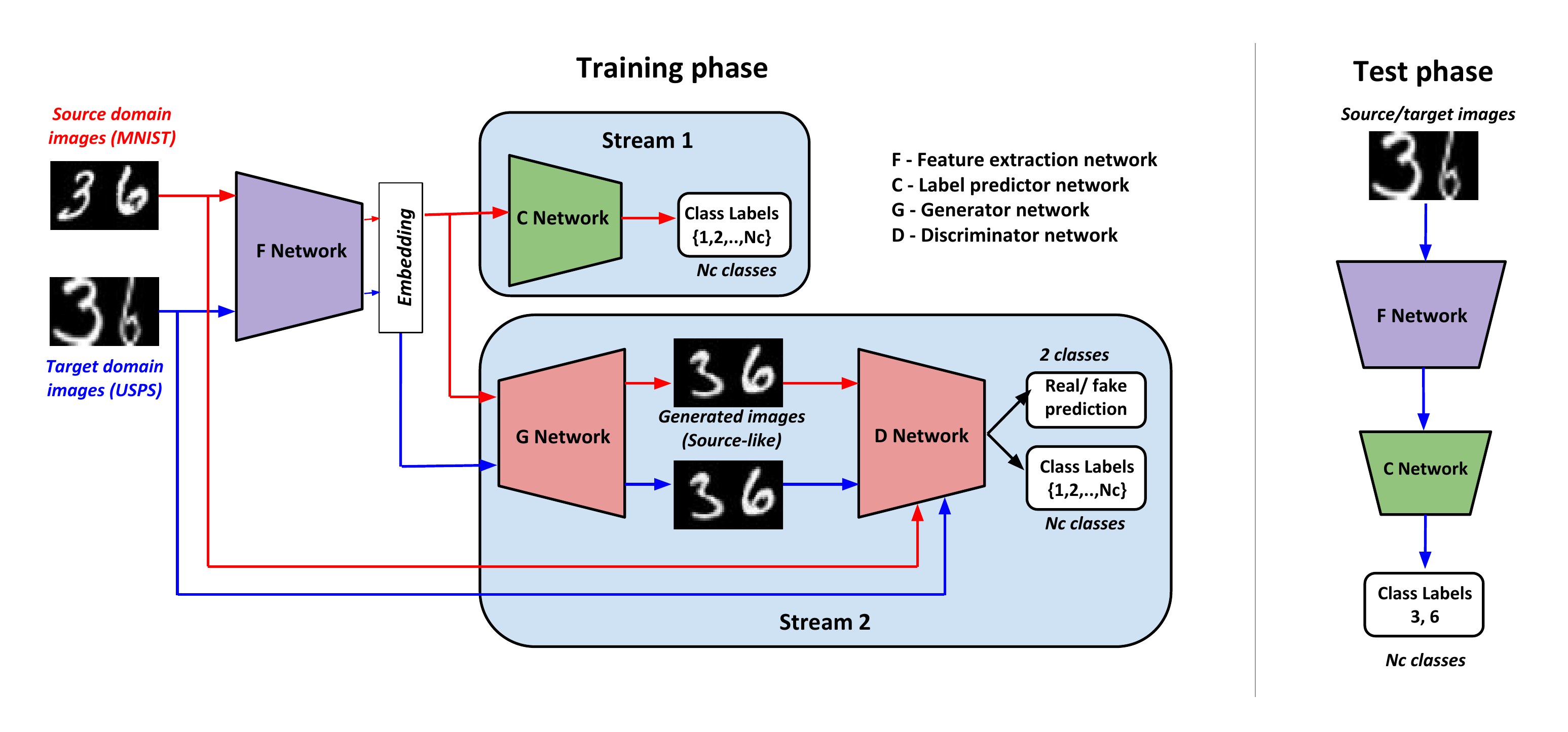}
\caption{Illustration of the proposed approach. In the training phase, our pipeline consists of two parallel streams - (1) Stream 1: classification branch where F-C networks are updated using supervised classification loss and (2) Stream 2: adversarial branch which is a Auxiliary Classifier GAN (ACGAN) framework (G-D pair). F-G-D networks are updated so that both source and target embeddings produce source-like images. Note: The auxiliary classifier in ACGAN uses only the source domain labels, and is needed to ensure that class-consistent images are generated (e.g) embedding of digit 3 generates an image that looks like 3. In the test phase, we remove Stream 2, and classification is performed using the F-C pair}
\label{fig:title}
\end{figure*}

The development of powerful learning algorithms such as Convolutional Neural Networks (CNNs) has provided an effective  pipeline for solving many classification problems \cite{cnnbaseline2014}. The abundance of labeled data has resulted in remarkable improvements for tasks such as the Imagenet challenge: beginning with the CNN framework of AlexNet \cite{imagenet} and more recently ResNets \cite{resnet2016} and its variants. Another example is the steady improvements in performance on  the LFW dataset \cite{facenet2015}. The common theme across all these approaches is the dependence on large amounts of labeled data. While labeled data is available and getting labeled data has been easier over the years, the lack of uniformity of label distributions across different domains results in suboptimal performance of even the most powerful CNN-based algorithms on realistic unseen test data. For example, labeled synthetic data is available in plenty but algorithms trained only on synthetic data perform poorly on real data. This is of vital importance in cases where labeled real data is unavailable. The use of such unlabeled target data to mitigate the shift between source and target distributions is the most useful  direction among domain adaptation approaches. Hence this paper focuses on the topic of unsupervised domain adaptation. 
In this work, we learn an embedding that is robust to the shift between source and target distributions. We achieve this by using unsupervised data sampled from the target distribution to guide the supervised learning procedure that uses data sampled from the source distribution. We propose an adversarial image generation approach to directly learn the shared feature embedding using labeled data from source and unlabeled data from the target. It should be noted that while there have been a few approaches that use an adversarial framework for solving the domain adaptation problem, the novelty of the proposed approach is in using a joint generative discriminative method: the embeddings are learned using a combination of classification loss and an image generation procedure that is modeled using a variant of Generative Adversarial Networks (GANs)~\cite{gan2014}. 

Figure \ref{fig:title} illustrates the pipeline of the proposed approach. During training, the source images are passed through the feature extraction network (encoder) to obtain an embedding which is then used by the label prediction network (classifier) for predicting the source label and also used by the generator to generate a realistic source image. The realistic nature of the images from the generator ($G$) is controlled by the discriminator ($D$). The encoder is updated based on the discriminative gradients from the classifier and generative gradients from the adversarial framework. Given unlabeled target images, the encoder is updated using only gradients from the adversarial part, since the labels are unavailable. Thus, the encoder learns to discriminate better even in the target domain using the knowledge imparted by the generator-discriminator pair. 
By using the discriminator as a multi-class classifier, we ensure that the gradient signals backpropagated by the discriminator for the unlabeled target images belong to the feature space of the respective classes. By sampling from the distribution of the generator after training, we show that the network has indeed learned to bring the source and target distributions closer. 

The main contribution of this work is to provide an adversarial image generation approach for unsupervised domain adaptation that directly learns a joint feature space in which the distance between source and target distributions is minimized. Different from contemporary approaches that achieve a similar objective by using a GAN as a data augmenter, our approach achieves superior results even in cases where a stand along image generation process is bound to fail (such as in the OFFICE dataset). This is done by utilizing the GAN framework to address the domain shift directly in the feature space learnt by the encoder. Our experiments show that the proposed approach yields superior results compared to similar approaches which update the embedding based on auto-encoders \cite{drcn2016} or disentangling the domain information from the embedding by learning a separate domain classifier \cite{ganin2014}. 

This paper is organized as follows: We begin in Section \ref{sec:background} by describing existing approaches for the unsupervised domain adaptation problem. In Section \ref{sec:method}, we describe in detail the formulation of our approach and the iterative training procedure. The experimental setups and the results are discussed in Section \ref{sec:results} using both quantitative and qualitative experiments, followed by discussion and conclusion in Section \ref{sec:conclusion}

\section{Related Work}
\label{sec:background}
Domain adaptation is an actively researched topic in many areas of Artificial Intelligence including Machine Learning, Natural Language Processing and Computer Vision. In this section, we describe techniques related to visual domain adaptation. Earlier approaches to domain adaptation focused on building feature representations that are invariant across domains. This was accomplished either by feature reweighting and selection mechanisms\cite{Huang_sampleselection} \cite{Daume_DA}, or by learning an explicit feature transformation that aligns source distribution to the target distribution (\cite{Gopalan_DA}, \cite{Pan_DA}, \cite{GFK}). The ability to deep neural networks to learn powerful representations [\cite{imagenet}, \cite{resnet2016}] has been harnessed to perform unsupervised domain adaptation in recent works [\cite{ganin2014}, \cite{DDC}, \cite{DAN},  \cite{RTN}, \cite{ADDA}]. The underlying idea behind such methods is to minimize a suitable loss function that captures domain discrepancy, in addition to the task being solved. 

Deep learning methods for visual domain adaptation can be broadly grouped into few major categories. One line of work uses Maximum Mean Discrepancy(MMD) as a metric to measure the shift across domains. Deep Domain Confusion (DDC)~\cite{DDC} jointly minimizes the classification loss and MMD loss of the last fully connected layer. Deep Adaptation Networks (DAN)~\cite{DAN} extends this idea by embedding all task specific layers in a reproducing kernel Hilbert space and minimizing the MMD in the projected space. In addition to MMD, Residual Transfer Networks (RTN)~\cite{RTN} uses a gated residual layer for classifier adaptation. Joint Adaptation Networks~\cite{JAN} learn a transfer network by aligning the joint distributions of multiple domain-specific layers across domains based on a Joint Maximum Mean Discrepancy (JMMD) criterion.

Another class of methods uses adversarial losses to perform domain adaptation. Revgrad~\cite{ganin2014} employs a domain classification network which aims to discriminate the source and the target embeddings. The goal of the feature extraction network is to produce embeddings that maximize the domain classifier loss, while at the same time minimizing the label prediction loss. This is accomplished by negating the gradients coming from the domain classification network. Adversarial Discriminative Domain Adaptation (ADDA)~\cite{ADDA} on the other hand learns separate feature extraction networks for source and target, and trains the target CNN so that a domain classifier cannot distinguish the embeddings produced by the source or target CNNs.

While methods discussed above apply adversarial losses in the embedding space, there has been a lot of interest recently to perform adaptation by applying adversarial losses in the pixel space. Such approaches primarily use generative models such as GANs to perform cross-domain image mapping. \cite{cross_domain_gan} and \cite{PixelDA} use adversarial networks to map source images to target and perform adaptation in the transferred space. Coupled GAN (CoGAN)~\cite{CoGAN} on the other hand trains a coupled generative model that learns the joint data distribution across the two domains. A domain invariant classifier is learnt by sharing weights with the discriminator of the CoGAN network. 

\textbf{Comparison to recent GAN-based DA approaches:}  While previous approaches such as \cite{cross_domain_gan} and \cite{PixelDA} use GANs as a data augmentation step, we use a GAN to obtain rich gradient information that makes the learned embeddings domain adaptive. Unlike the previous methods, our approach does not completely rely on a successful image generation process. As a result, our method works well in cases where image generation is hard (eg. in the OFFICE dataset where the number of samples per class is limited). We observed that in such cases, even though the generator network we use performs a mere style transfer, yet this is sufficient for providing good gradient information for successfully aligning the domains, as demonstrated by our superior performance on the OFFICE dataset. 

\section{Approach}\label{sec:method}
\paragraph{Problem Description:} In this section, we provide a formal treatment of the proposed approach and discuss in detail our iterative optimization procedure. Let $\mathbf{X}=\{x_i\}_{i=1}^{N}$ be an input space of images and $\mathbf{Y} = \{y_i\}_{i=1}^{N}$ be the label space. We assume there exists a source distribution, $\mathcal{S}(x,y)$ and target distribution $\mathcal{T}(x,y)$ over the samples in $\mathbf{X}$. In unsupervised domain adaptation, we have access to the source distribution using labeled data from $\mathbf{X}$ and the target distribution via unlabeled data sampled from $\mathbf{X}$. Operationally, the problem of unsupervised domain adaptation can be stated as learning a predictor that is optimal in the joint distribution space by using labeled source data and unlabeled target data sampled from $\mathbf{X}$. We consider problems where the data from $\mathbf{X}$ takes discrete labels from the set $\mathbf{L}=\{1,2,3,...N_c\}$, where $N_c$ is the total number of classes. Our objective is to learn an embedding map $F:\mathbf{X}\mapsto\mathbb{R}^d$ and a prediction function $C:\mathbb{R}^d\mapsto\mathbf{L}$. In this work, both $F$ and $C$ are modeled as deep neural networks. The predictor has access to the labels only for the data sampled from the source distribution and not from the target distribution. By extracting information from the target data during training, $F$ implicitly learns the domain shift between $\mathcal{S}$ and $\mathcal{T}$. In the rest of this section, we use the terms source (target) distribution and source (target) domain interchangeably.

Several approaches including learning entropy-based metrics \cite{RTN}, learning a domain classifier based on a embedding network \cite{ganin2014} or denoising autoencoders \cite{drcn2016} have been used to transfer information between source and target distributions. In this work, we propose a GAN-based approach to bridge the gap between source and target domains. We accomplish this by using both generative and a discriminative processes thus ensuring a rich information transfer to the learnt embedding.

\textbf{Overview of GANs:} In a traditional GAN, two competing mappings are learned: the discriminator $D$ and the generator $G$, both of which are modeled as deep neural networks. $G$ and $D$ play a minmax game where $D$ tries to classify the generated samples as fake and $G$ tries to fool $D$ by producing examples that are as realistic as possible. More formally, to train a GAN, the following optimization problem is solved in an iterative manner:
 \begin{equation}
 \begin{aligned}
   \mymin[G] \mymax[D] \, & \mathbf{E}_{x\sim p_{data}} (\log(D(x)) \\
   & + \mathbf{E}_{z\sim p_{noise}} \log(1-D(G(z)))
 \end{aligned}
 \end{equation}
 $D(x)$ represents the probability that $x$ came from the real data distribution rather than the distribution modeled by the generator $G$. As an extension to traditional GANs, conditional GANs \cite{conditionalgan2014} enable conditioning the generator and discriminator mappings on additional data such as a class label or an embedding. They have been shown to generate images of digits and faces conditioned on the class label or the embedding respectively~\cite{cross_domain_gan}. Training a conditional GAN  involves optimizing the following minimax objective: 
 \begin{equation}
 \begin{aligned}
  \mymin[G] \mymax[D] \, & \mathbf{E}_{x\sim p_{data}} (\log(D(x|y)) \\ 
  & + \mathbf{E}_{\{z\sim p_{noise}\}} \log(1-D(G(z|y)))
  \end{aligned}
 \end{equation}
 
\paragraph{Proposed Approach:} In this work, we employ a variant of the conditional GAN called Auxiliary Classifier GAN (AC-GAN) \cite{acgan2016} where the discriminator is modeled as a multi-class classifier instead of providing conditioning information at the input. We modify the AC-GAN set up for the domain adaptation problem as follows: 

(a) Given a real image $x$ as input to $F$, the input to the generator network $G$ is $x_{g}=[F(x),z,l]$, which is a concatenated version of the encoder embedding $F(x)$, a random noise vector $z \in \mathbb{R}^{d}$ sampled from $\mathcal{N}(0,1)$ and a one hot encoding of the class label, $l \in \{0,1\}^{(N_c+1)}$ with $N_c$ real classes and $\{N_c+1\}$ being the fake class. For all target samples, since the class labels are unknown, $l$ is set as the one hot encoding of the fake class $\{N_c+1\}$. 

(b) We employ a classifier network $C$ that takes as input the embedding generated by $F$ and predicts a multiclass distribution $C(\mathrm{x})$ i.e. the class probability distribution of the input $\mathrm{x}$, which is modeled as a ($N_c$)-way classifier.

(c) The discriminator mapping $D$ takes the real image $x$ or the generated image $G(x_{g})$ as input and outputs two distributions: (1) $D_{data}(x)$: the probability of the input being real, which is modeled as a binary classifier. (2) $D_{cls}(x)$: the class probability distribution of the input $x$, which is modeled as a ($N_c$)-way classifier. To clarify notation, we use ${D_{cls}(x)}_{y}$ to imply the probability assigned by the classifier mapping $D_{cls}$ for input $x$ to class $y$. It should be noted that, for target data, since class labels are unknown, only $D_{data}$ is used to backpropagate the gradients. 

\begin{algorithm*}[!htb]
\caption{Iterative training procedure of our approach}\label{alg:algo}
\label{tab:algo}
\begin{algorithmic}[1]
\State training iterations = N
\For{t in 1:N}
		\State Sample $k$ images with labels from source domain $\mathcal{S}$: $\{s_i,y_i\}_{i=1}^{k}$
        \State Let $f_i = F(s_i)$ be the embeddings computed for the source images.
        \State Sample $k$ images from target domain $\mathcal{T}$ : $\{t_i\}_{i=1}^{k}$
        \State Let $h_i = F(t_i)$ be the embeddings computed for the target images.        
		\State Sample $k$ random noise samples $\{z_i\}_{i=1}^{k} \sim \mathcal{N}(0,1)$.	
        \State Let $f_{g_{i}}$ and $h_{g_{i}}$ be the concatenated inputs to the generator.
        \State Update discriminator using the following objectives:
        \begin{equation}
         L_{D} = L_{data,src} + L_{cls,src} + L_{adv,tgt}
         \end{equation}         
        \begin{itemize}
        \setlength{\itemindent}{.25in}
        \item $L_{data,src} = \mymax[D] \frac{1}{k}\sum_{i=1}^{k} \log(D_{data}(s_i)) + log(1-D_{data}(G(f_{g_{i}})))$
        \item $L_{cls,src} = \mymax[D] \frac{1}{k}\sum_{i=1}^{k} \log({D_{cls}(s_i)}_{y_i})$
        \item $L_{adv,tgt} = \mymax[D] \frac{1}{k}\sum_{i=1}^{k} \log(1-D_{data}(G(h_{g_{i}})))$
        \end{itemize}
             
         \State Update the generator, only for source data, through the discriminator gradients computed using real labels.
        \begin{equation}
        L_{G} = \mymin[G] \frac{1}{k}\sum_{i=1}^{k} - \log(D_{cls}(G(f_{g_i}))_{y_i}) + \log(1-D_{data}(G(f_{g_i}))) 
        \end{equation}
         \State Update the embedding $F$ using a linear combination of the adversarial loss and classification loss.  Update the classifier $C$ for the source data using a cross entropy loss function.
        \begin{equation}
        L_{F} =  L_{C} +  \alpha\,L_{cls,src} + \beta\,L_{F_{adv}}
        \end{equation}
        \begin{itemize}
        \setlength{\itemindent}{.25in}
        \item $L_{C} = \mymin[C] \mymin[F] \frac{1}{k}\sum_{i=1}^{k} - \log(C(f_i)_{y_i})$
        \item $L_{cls,src} = \mymin[F] \frac{1}{k} \sum_{i=1}^{k} - \log(D_{cls}(G(f_{g_i}))_{y_i}) $
        \item $L_{F_{adv}} = \mymin[F] \frac{1}{k} \sum_{i=1}^{k}  \log(1-D_{data}(G(h_{g_i})))$        
        \end{itemize}       
  
\EndFor
\end{algorithmic}
\end{algorithm*}
 
Now, we describe our optimization procedure in detail. To jointly learn the embedding and the generator-discriminator pair, we optimize the $D$, $G$, $F$ and $C$ networks in an alternating manner:
\begin{enumerate}
\item Given source images as input, $D$ outputs two distributions $D_{data}$ and $D_{cls}$. $D_{data}$ is optimized by minimizing a binary cross entropy loss $L_{data,src}$ and $D_{cls}$ is optimized by minimizing the cross entropy loss $L_{cls,src}$ between the source labels and the model predictive distribution $D_{cls}(x)$. In the case of source inputs, the gradients are generated using the following loss functions:
\begin{equation}
\begin{aligned}
& L_{data,src} + L_{cls,src} = \mathbf{E}_{x\sim \mathcal{S}} \mymax[D] \log(D_{data}(x))   \\
& + \log(1-D_{data}(G(x_g))) + \log({D_{cls}(x)}_{y}) \\
\end{aligned}
\end{equation}
\item Using the gradients from $D$, $G$ is updated using a combination of adversarial loss and classification loss to produce realistic class consistent source images.
\begin{equation}
\begin{aligned}
& L_{G} = \mymin[G] \mathbf{E}_{x\sim \mathcal{S}} - \log({D_{cls}(G(x_{g}))}_{y}) \\
& \quad + \log(1-D_{data}(G(x_g))) ,\\
\end{aligned}
\label{eq:gen_update}
\end{equation}

\item $F$ and $C$ are updated based on the source images and source labels in a traditional supervised manner. $F$ is also updated using the adversarial gradients from $D$ so that the feature learning and image generation processes co-occur smoothly. 
\begin{equation}
\begin{aligned}
& L_{C} = \mymin[C] \mymin[F] \mathbf{E}_{x\sim \mathcal{S}} - \log(C(F(x))_{y}) ,\\
& L_{cls,src}  = \mymin[F] \mathbf{E}_{x\sim \mathcal{S}} - \alpha \, \log({D_{cls}(G(x_g))}_{y}))\\
\end{aligned}
\end{equation}
\item In the final step, the real target images are presented as input to $F$. The target embeddings output by $F$ along with the random noise vector $z$ and the fake label encoding $l$ are input to $G$. The generated target images $G(x_g)$ are then given as input to $D$. As described above, $D$ outputs two distributions but the loss function is evaluated only for $D_{data}$ since in the unsupervised case considered here, target labels are not provided during training. Hence, $D$ is updated to classify the generated target images as fake as follows:
\begin{equation}
\begin{aligned}
& L_{adv,tgt} = \mymax[D] \mathbf{E}_{x\sim \mathcal{T}} \log(1-D_{data}(G(x_g))) \\
\end{aligned}
\end{equation}

In order to transfer the knowledge of target distribution to the embedding, $F$ is updated using the gradients from $D_{data}$ that corresponds to the generated target images being classified as real: 
\begin{equation}
\begin{aligned}
& L_{F_{adv}} = \mymin[F] \mathbf{E}_{x\sim \mathcal{T}} \, \beta \, \log(1-D_{data}(G(x_g))) \\
\end{aligned}
\end{equation}

\end{enumerate}


The proposed iterative optimization procedure is summarized as a pseudocode in Algorithm~\ref{tab:algo}. $\alpha$ and $\beta$ are the coefficients that trade off between the classification loss and the source and target adversarial losses. Based on our experiments, we find that our approach is not overly sensitive to the cost coefficients $\alpha$ and $\beta$. However, the value of the parameter is dependent on the application and size of the dataset. Such specifications are mentioned in the supplementary material.

\paragraph{Use of unlabeled target data:}
The main strength of our approach is how the target images are used to update the embedding. Given a batch of target images as input, we update the embedding $F$ by using the following binary loss term:
\begin{equation}
\mymin[F] \beta \, \log(1-D_{data}(G(x_{g}))
\label{target_update}
\end{equation}

where $x_g$ is the concatenated input to $G$ as described earlier and $\beta$ is the weight coefficient for the target adversarial loss. The use of target data is intended to bring the source and target distributions closer in the feature space learned by $F$. To achieve this, we update the $F$ network to produce class consistent embeddings for both source and target data. Performing this update for source data is straightforward since the source labels are available during training. Since labels are unavailable for target data, we use the generative ability of the $G$-$D$ pair for obtaining the required gradients. 

Given source inputs, $G$ is updated to fool $D$ using gradients from Eq. (\ref{eq:gen_update}) which provide the conditioning required for $G$ to produce class consistent fake images. Given target inputs, the update in Eq. (\ref{target_update}) encourages $F$ to produce target embeddings that are aligned with the source distribution. As training progresses, the class conditioning information learned by $G$ during the source update (Eq. (\ref{eq:gen_update})) was found to be sufficient for it to produce class consistent images for target embeddings as well. This symbiotic relationship between the embedding and the adversarial framework contributes to the success of the proposed approach.

\section{Experiments and Results}\label{sec:results}

This  section reports the experimental validation of our approach. We perform a thorough study by conducting experiments across three adaptation settings: (1) low domain shift and simple data distribution: DIGITS dataset, (2) moderate domain shift and complex data distribution: OFFICE dataset, (3) high domain shift and complex data distribution: Synthetic to real adaptation. By complex data distribution, we denote datasets containing images with high variability and limited number of samples. Our methods performs well in all three regimes, thus demonstrating the versatility of our approach. \footnotetext{Training code: \url{https://goo.gl/zUVeqC}}



\begin{table*}[t]
\centering
\caption{Accuracy (mean $\pm$ std\%) values for cross-domain recognition tasks over five independent runs on the digits based datasets. The best numbers are indicated in \textbf{bold} and the second best are \underline{underlined}. $-$ denotes unreported results. MN: MNIST,   US: USPS,   SV: SVHN. MN$\to$US (p) denotes the MN$\to$US experiment run using the protocol established in ~\cite{transfer_feat_learning}, while MN$\to$US (f) denotes the experiment run using the entire datasets. (Refer to Digits experiments section for more details)}
\label{tab:digits_table}
\begin{tabular}{|c|c|c|c|c|c|}
\hline 
\rule{0pt}{3ex}
Method & MN $\rightarrow$ US (p) & MN $\rightarrow$ US (f) & US $\rightarrow$ MN & SV $\rightarrow$ MN  \\
\hline
\rule{0pt}{3ex}
Source only & 75.2 $\pm$ 1.6 & 79.1 $\pm$ 0.9 & 57.1 $\pm$ 1.7 & 60.3 $\pm$ 1.5  \\
RevGrad~\cite{ganin2014}& 77.1 $\pm$ 1.8 & - & 73.0 $\pm$ 2.0 & 73.9  \\
DRCN~\cite{drcn2016}& \underline{91.8} $\pm$ 0.09 & - & 73.7 $\pm$ 0.04 & \underline{82.0} $\pm$ 0.16  \\
CoGAN~\cite{CoGAN}& 91.2 $\pm$ 0.8 & - & 89.1 $\pm$ 0.8 & -   \\
ADDA~\cite{ADDA}& 89.4 $\pm$ 0.2 & - & \underline{90.1} $\pm$ 0.8 & 76.0 $\pm$ 1.8  \\
PixelDA~\cite{PixelDA} & - & \textbf{95.9} & - & -  \\
Ours &  \textbf{92.8 $\pm$ 0.9} & \underline{95.3} $\pm$ 0.7 & \textbf{90.8} $\pm$ 1.3 & \textbf{92.4} $\pm$ 0.9  \\

\hline
\end{tabular}
\\
\rule{0pt}{3ex}
\end{table*}

\begin{table*}[t]
\centering
\caption{Accuracy (mean $\pm$ std\%) values on the OFFICE dataset for the standard protocol for unsupervised domain adaptation~\cite{GFK}. Results are reported as an average over 5 independent runs. The best numbers are indicated in \textbf{bold} and the second best are \underline{underlined}. $-$ denotes unreported results. A: Amazon, W: Webcam, D: DSLR}
\label{tab:office_table}
\resizebox{\textwidth}{!}{
\begin{tabular}{|c|c|c|c|c|c|c|c|}
\hline 
\rule{0pt}{3ex}
Method & A $\rightarrow$ W & D $\rightarrow$ W & W $\rightarrow$ D & A $\rightarrow$ D & D $\rightarrow$ A & W $\rightarrow$ A & Average\\
\hline
\rule{0pt}{3ex}
ResNet - Source only~\cite{resnet2016} & 68.4 $\pm$ 0.2 & 96.7 $\pm$ 0.1 & 99.3 $\pm$ 0.1 & 68.9 $\pm$ 0.2 & 62.5 $\pm$ 0.3 & 60.7 $\pm$ 0.3 & 76.1 \\
TCA~\cite{Pan_DA}& 72.7 $\pm$ 0.0 & 96.7 $\pm$ 0.0 & \underline{99.6} $\pm$ 0.0 & 74.1 $\pm$ 0.0 & 61.7 $\pm$ 0.0 & 60.9 $\pm$ 0.0 & 77.6 \\
GFK~\cite{GFK} & 72.8 $\pm$ 0.0 & 95.0 $\pm$ 0.0 & 98.2 $\pm$ 0.0 & 74.5 $\pm$ 0.0 & 63.4 $\pm$ 0.0 & 61.0 $\pm$ 0.0 & 77.5 \\
DDC~\cite{DDC} & 75.6 $\pm$ 0.2 & 76.0$\pm$ 0.2 & 98.2 $\pm$ 0.1 & 76.5 $\pm$ 0.3 & 62.2 $\pm$ 0.4 & 61.5 $\pm$ 0.5 & 78.3 \\
DAN	~\cite{DAN}& 80.5 $\pm$ 0.4 & 97.1 $\pm$ 0.2 & \underline{99.6} $\pm$ 0.1 & 78.6 $\pm$ 0.2 & 63.6 $\pm$ 0.3 & 62.8 $\pm$ 0.2 & 80.4 \\
RTN~\cite{RTN} & 84.5 $\pm$ 0.2 & 96.8 $\pm$ 0.1 & 99.4 $\pm$ 0.1 & 77.5 $\pm$ 0.3 & 66.2 $\pm$ 0.2 & 64.8 $\pm$ 0.3 & 81.6 \\
RevGrad~\cite{ganin2014} & 82.0 $\pm$ 0.4 & 96.9 $\pm$ 0.2 & 99.1 $\pm$ 0.1 & 79.4 $\pm$ 0.4 & 68.2 $\pm$ 0.4 & 67.4 $\pm$ 0.5 & 82.2 \\
JAN~\cite{JAN} & \underline{85.4} $\pm$ 0.3 & \underline{97.4} $\pm$ 0.2 & \textbf{99.8} $\pm$ 0.2 & \underline{84.7} $\pm$ 0.3 & \underline{68.6} $\pm$ 0.3 & \underline{70.0} $\pm$ 0.4 & \underline{84.3} \\
Ours		& \textbf{89.5} $\pm$ 0.5 & \textbf{97.9}  $\pm$ 0.3& \textbf{99.8} $\pm$ 0.4 & \textbf{87.7} $\pm$ 0.5 & \textbf{72.8}  $\pm$ 0.3 & \textbf{71.4} $\pm$ 0.4 & \textbf{86.5} \\
\hline
\end{tabular}
}
\end{table*}
\subsection{Digit Experiments}
The first set of experiments involve digit classification in three standard DIGITS datasets: MNIST~\cite{mnist}, USPS~\cite{usps} and SVHN~\cite{svhn}. Each dataset contains digits belonging to $10$ classes (0-9). MNIST and USPS are large datasets of handwritten digits captured under constrained conditions. SVHN dataset, on the other hand was obtained by cropping house numbers in Google Street View images and hence captures much more diversity. We test the three common domain adaptation settings: SVHN $\to$ MNIST, MNIST $\to$ USPS and USPS $\to$ MNIST. In each setting, we use the label information only from the source domain, thus following the unsupervised protocol. 

For all digit experiments, following other recent works ~\cite{ganin2014}\cite{ADDA}, we use a modified version of Lenet architecture as our $F$ network. For $G$ and $D$ networks, we use architectures similar to those used in DCGAN~\cite{dcgan2015}. 

\subsubsection*{(a) MNIST $\leftrightarrow$ USPS}
\vspace{-2mm}
We start with the easy case of adaptation involving MNIST and USPS. The MNIST dataset is split into $60000$ training and $10000$ test images, while the USPS dataset contains $7291$ training and $2007$ test images. We run our experiments in two settings: (1) using the entire training set of MNIST and USPS (MNIST $\leftrightarrow$USPS (f)), and (2) using the protocol established in \cite{transfer_feat_learning}, sampling $2000$ images from MNIST and $1800$ images from USPS (MNIST $\leftrightarrow$USPS (p)). Table.~\ref{tab:digits_table} presents the results of the proposed approach in comparison with other contemporary approaches. The reported numbers are averaged over 5 independent runs with different random samplings or initializations. We can observe that our approach achieves the best performance in all cases except in the MNIST $\to$ USPS full protocol case where our accuracy is very close to the best performing method.
\vspace{-2mm}
\subsubsection*{(b) SVHN $\rightarrow$ MNIST}
\begin{figure}[!h]
\begin{subfigure}{.25\textwidth}
  \centering
  \includegraphics[width=\linewidth]{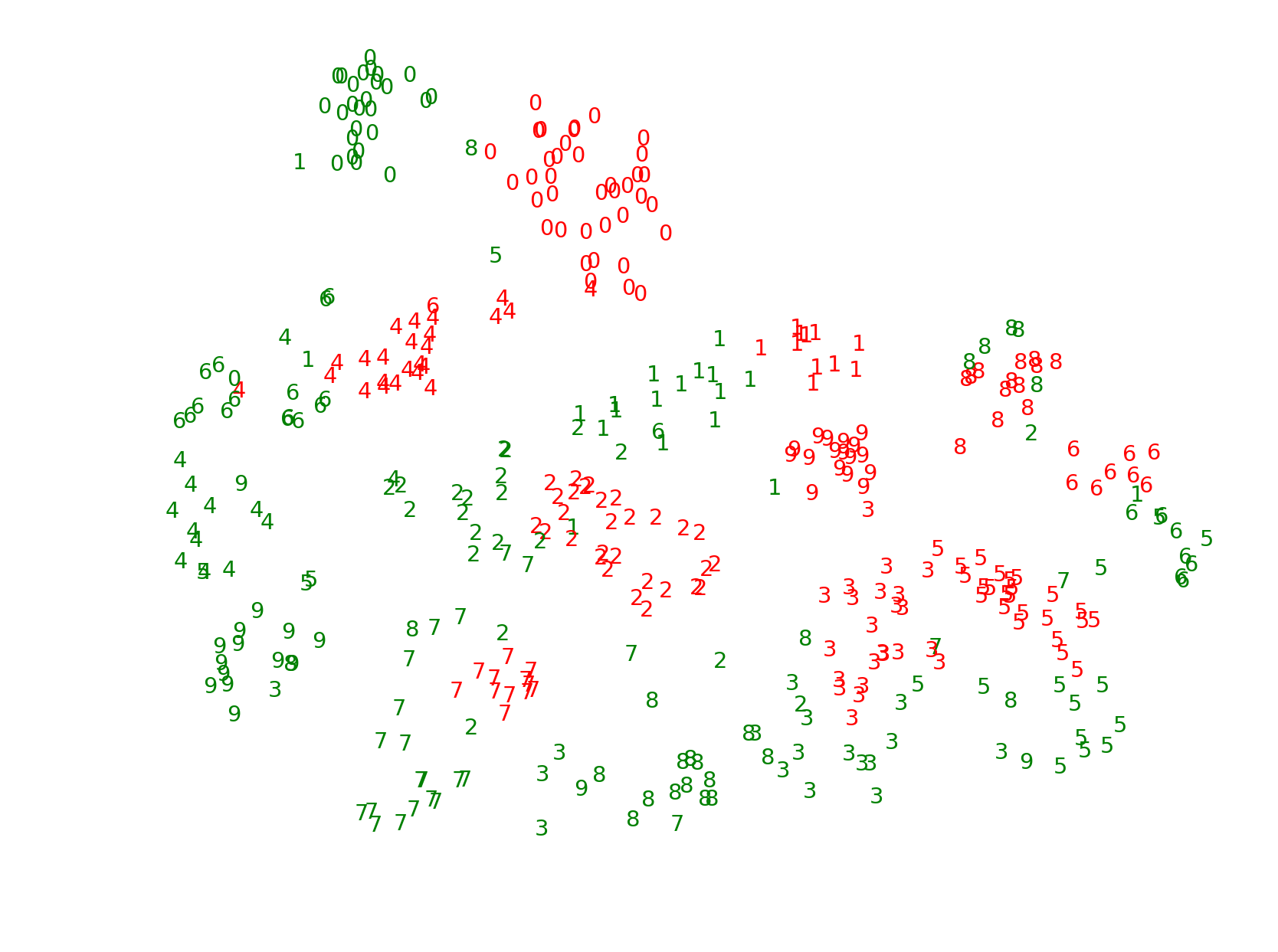}
  \caption{Non adapted}
  \label{fig:sub1}
\end{subfigure}%
\begin{subfigure}{.25\textwidth}
  \centering
  \includegraphics[width=\linewidth]{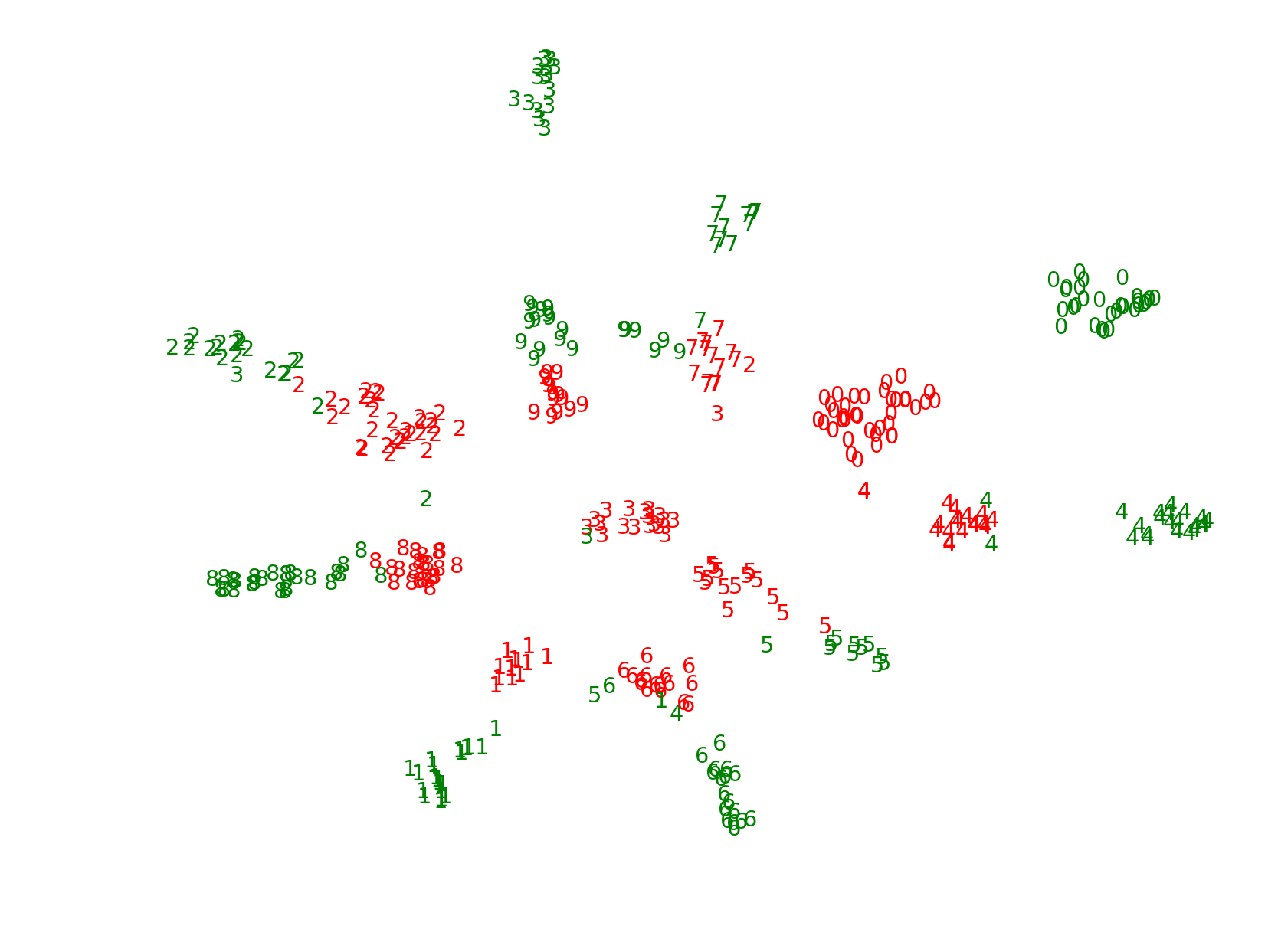}
  \caption{Adapted}
  \label{fig:sub2}
\end{subfigure}
\caption{TSNE visualization of SVHN $\rightarrow$ MNIST adaptation. In (a), the source data shown in \textit{red} is classified well into distinct clusters but the target data is clustered poorly. On applying the proposed approach, as shown in  (b), both the source and target distributions are brought closer in a class consistent manner.}
\label{fig:tsne}
\end{figure}

Compared to the previous experiment, SVHN $\rightarrow$ MNIST presents a harder case of domain adaptation owing to larger domain gap. Following other works~\cite{ganin2014} ~\cite{ADDA}, we use the entire training set (labeled $73257$ SVHN images and unlabeled $60000$ MNIST images) to train our model, and evaluate on the training set of the target domain (MNIST dataset). From Table.~\ref{tab:office_table}, we observe that our method significantly improves the performance of the source-only model from $60.3\%$ to $92.4\%$, which results in a performance gain of $32.1\%$. We also outperform other methods by a large margin, obtaining at least $10.4\%$ performance improvement. A visualization of this improvement in performance is done in figure \ref{fig:tsne}, where we show a t-SNE plot of the features of the embedding network $F$ for the adapted and non-adapted cases. 

\subsection{OFFICE experiments}
The next set of experiments involve the OFFICE dataset, which is a small scale dataset containing images belonging to $31$ classes from three domains - Amazon, Webcam and DSLR, each containing 2817, 795 and  498 images respectively. The small dataset size poses a challenge to our approach since we rely on GAN which demands more data for better image generation. Nevertheless, we perform experiments on the OFFICE dataset to demonstrate that though our method does not succeed in generating very realistic images, the approach still results in improved performance by using the generative process to obtain domain invariant feature representations.


Training deep networks with randomly initialized weights on small datasets give poor performance. So, an effective technique used in practice is to fine-tune networks trained on a related task having large data~\cite{how_transferrable}. Following this rationale, we initialized the $F$ network using a pre-trained ResNet-50~\cite{resnet2016} model trained on Imagenet. For $D$ and $G$ networks, we used architectures similar to the ones used in the Digits experiments. It should be noted that even though the inputs are $224 \times 224$, the $G$ network is made to generate a downsampled version of size $64 \times 64$. Standard data augmentation steps involving mean normalization, random cropping and mirroring were performed. 

In all our experiments, we follow the standard unsupervised protocol - using the entire labeled data in the source domain and unlabeled data in the target domain. Table~\ref{tab:office_table} reports the performance of our method in comparison to other methods. We observe that our method obtains the state-of-the-art performance in all the settings. In particular, we get good performance improvement consistently in all hard transfer cases: $A \rightarrow W$, $A \rightarrow D$, $W \rightarrow A$ and $D \rightarrow A$. 


\subsection{Synthetic to Real experiments}

To test the effectiveness of the proposed approach further, we perform experiments in the hardest case of domain adaptation involving adaptation from synthetic to real datasets. This setting is particularly interesting because of its enormous practical implications. In this experiment, we use CAD synthetic dataset~\cite{peng15iccv} and a subset of PASCAL VOC dataset~\cite{PASCAL} as our source and target sets respectively. The CAD synthetic dataset contains multiple renderings of 3D CAD models of the $20$ object categories contained in the PASCAL dataset. To create the datasets, we follow the protocol described in ~\cite{synth2real_dataset}: The CAD dataset contains six subsets with different configurations (i.e. RR-RR, W-RR, W-UG, RR-UG, RG-UG, RG-RR).  Of these, we use images with white background (W-UG subset) as our training set. To generate the target set, we crop 14976 patches from 4952 images of the PASCAL VOC 2007 test set using the object bounding boxes provided. The lack of realistic background and texture in the CAD synthetic dataset increases the disparity from the natural image manifold, thus making domain adaptation extremely challenging.

Due to the high domain gap, we observed that models trained on the CAD synthetic dataset with randomly initialized weights performed very poorly on the target dataset. So, similar to the previous set of experiments, we initialized the $F$ network with pretrained models. In particular, we removed the last fully connected layer from the VGG16 model trained on Imagenet and used it as our $F$ network. Note that the same $F$ network is used to train all other methods for fair comparison. Table.~\ref{tab:syn2real_table} reports the results of the experiments we ran on the Synthetic to real setting. We can observe that our method improves the baseline performance from $38.1\%$ to $50.4\%$ in addition to outperforming all other compared methods.

\begin{table}[!h]
\centering
\caption{Accuracy (mean $\pm$ std\%) values over five independent runs on the Synthetic to real setting. The best numbers are indicated in \textbf{bold}.}
\label{tab:syn2real_table}
\resizebox{0.4\textwidth}{!}{%
\begin{tabular}{c|c}
\hline
\rule{0pt}{3ex}
Method & CAD $\rightarrow$ PASCAL \\
\hline
\hline 
\rule{0pt}{3ex}
VGGNet - Source only & 38.1 $\pm$ 0.4 \\
RevGrad~\cite{ganin2014}& 48.3 $\pm$ 0.7  \\
RTN~\cite{RTN} & 43.2 $\pm$ 0.5 \\
JAN~\cite{JAN} & 46.4 $\pm$ 0.8 \\
Ours &  \textbf{50.4} $\pm$ 0.6 \\
\hline
\end{tabular}
}
\end{table}

\subsection{VISDA challenge}

In this section, we present the results on VISDA dataset~\cite{visda_challenge} - a large scale testbed for unsupervised domain adaptation algorithms. The task is to train classification models on synthetic dataset generated from the renderings of 3D CAD models and adapt these models to real images which are drawn from Microsoft COCO~\cite{COCO}(validation set) and Youtube Bounding Box dataset~\cite{Youtube_BB}(test set). We train our models using the same hyper-parameter settings and data augmentation scheme as the previous experiment. Table.~\ref{tab:visda_results} presents the results on the VISDA classification challenge. We find that our method achieves significant performance gains compared to the baseline model.

\begin{table}
\centering
\caption{Performance (accuracy) of our approach on VISDA classification dataset}
\label{tab:visda_results}
\begin{tabular}{c|l|l|l}
\hline
\rule{0pt}{3ex}
\multirow{2}{*}{Model} & \multicolumn{3}{c}{Visda-C: \textit{Val}}  \\ \cline{2-4} \rule{0pt}{3ex}
                       & Source-only       & Adapted    & Gain \\ \hline
\hline
\rule{0pt}{3ex}
Resnet-18              & 35.3              & 63.1       &  78.7\% \\
Resnet-50              & 40.2              & 69.5       &  72.8\% \\ 
Resnet-152             & 44.5              & 77.1       &  73.2\% \\ 
\hline
\rule{0pt}{3ex}
  &  \multicolumn{3}{c}{Visda-C: \textit{Test}} \\ \hline
\hline
\rule{0pt}{3ex}
Resnet-152             & 40.9              & 72.3       &  76.7\% \\ 
\hline
\end{tabular}
\end{table}

\subsection{Ablation Study}

In this experiment, we study the effect of each individual component to the overall performance. The embedding network $F$ is updated using a combination of losses from two streams (1) supervised classification stream and (2) adversarial stream, as shown in Figure \ref{fig:title}. The adversarial stream consists of the G-D pair, with D containing two components - real/fake classifier which we denote as $C_{1}$, and auxiliary classifier which we denote as $C_{2}$. We report the performance on the following three settings: (1) using only the Stream 1 and only using source data to train - this corresponds to the Source-only setting (2) Using stream 1 + $C_{1}$ classifier from stream 2 - this corresponds to the case where source and target embeddings are forced to produce source-like images, but class information is not provided to the discriminator and (3) Using stream 1 + stream2 ($C_{1}$ + $C_{2}$) - this is our entire system. For settings (2) and (3) we utilized labeled source data and unlabeled target data during training. Table~\ref{tab:ablation} presents the results of this experiment. 
\vspace{-2mm}
\begin{table}[!h]
\centering
\caption{Ablation study for OFFICE A$\rightarrow$W setting}
\label{tab:ablation}
\resizebox{0.45\textwidth}{!}{
\begin{tabular}{c|c}
\hline
\rule{0pt}{3ex}
Setting & Accuracy(in $\%$) \\
\hline
\hline 
\rule{0pt}{3ex}
Stream 1 - Source only & 68.4 \\
Stream 1 + Stream 2 ($C_{1}$ only) & 80.5  \\
Stream 1 + Stream 2 ($C_{1}+C_{2}$) & 89.5 \\
\hline
\end{tabular}
}
\end{table}

We observe that using only the real/fake classifier $C_{1}$ in the discriminator does improve performance, but the auxiliary classifier $C_{2}$ is needed to get the full performance benefit. This can be attributed to the mode collapse problem in traditional GANs (we observed that training without $C_{2}$ resulted in missing modes and mismatched mappings where embeddings get mapped to images of wrong classes), hence resulting in sub-optimal performance. Use of an auxiliary classifier objective in $D$ stabilizes the GAN training as observed in \cite{acgan2016} and significantly improves the performance of our approach.
\section{Conclusion and Future Work}\label{sec:conclusion}
In this paper, we addressed the problem of unsupervised visual domain adaptation. We proposed a joint adversarial-discriminative approach that transfers the information of the target distribution to the learned embedding using a generator-discriminator pair. We demonstrated the superiority of our approach over existing methods that address this problem using experiments on three different tasks, thus making our approach more generally applicable and versatile. Some avenues for future work include using stronger encoder architectures and applications of our approach to more challenging domain adaptation problems such as RGB-D object recognition and  medical imaging.

\section*{Acknowledgement}
This research is based upon work supported by the Office of the Director of National Intelligence (ODNI), Intelligence Advanced Research Projects Activity (IARPA), via IARPA R\&D Contract No. 2014-14071600012.  The views and conclusions contained herein are those of the authors and should not be interpreted as necessarily representing the official policies or endorsements, either expressed or implied, of the ODNI, IARPA, or the U.S. Government.  The U.S. Government is authorized to reproduce and distribute reprints for Governmental purposes notwithstanding any copyright annotation thereon.







 
\section{Network Architectures and Hyperparameters}
\begin{figure}
\centering
\includegraphics[width=0.5\textwidth,height=1.5\linewidth]{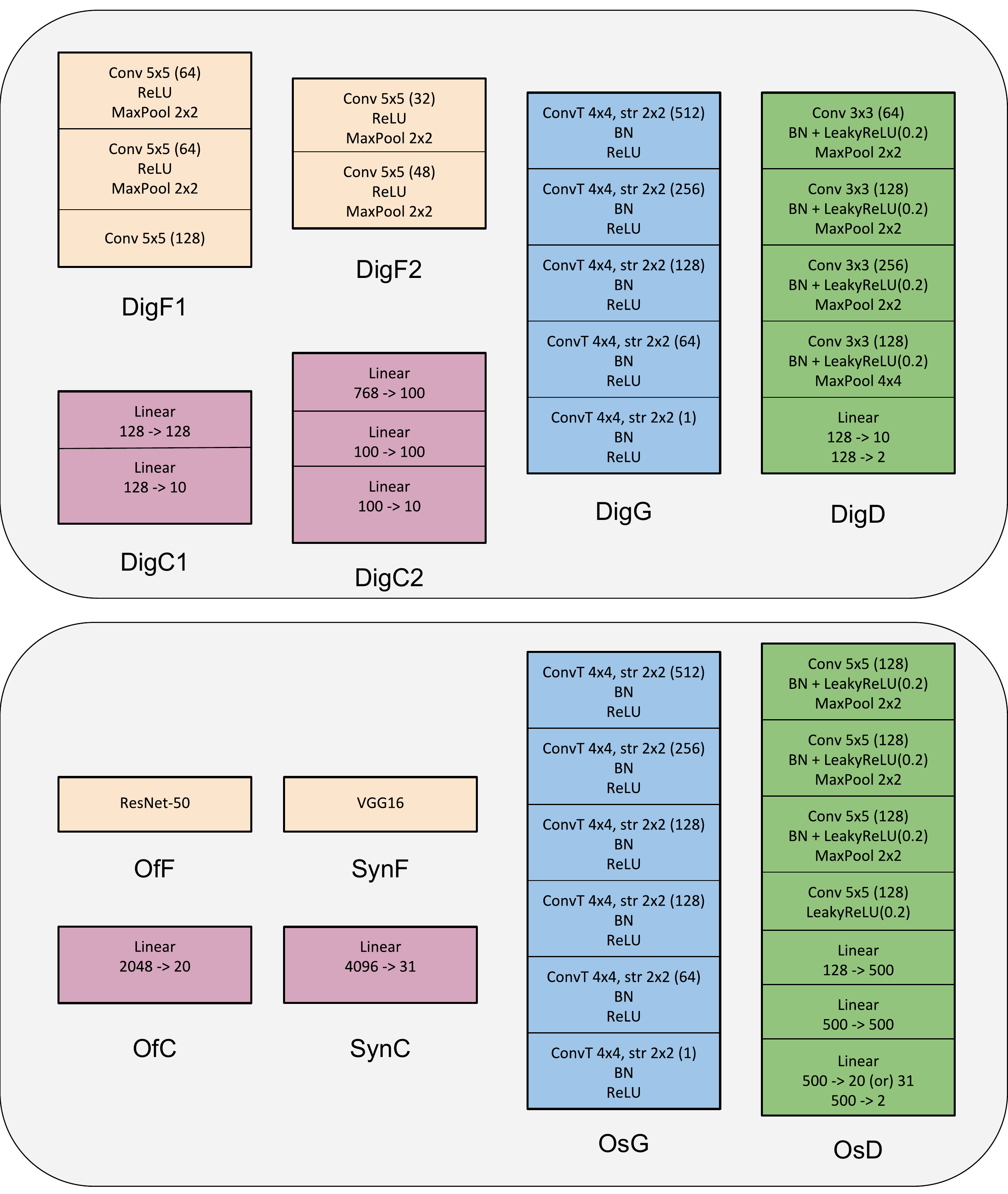}
\caption{Network Architectures. Legend: BN - Batch Normalization, ConvT - Transposed convolution layer}
\label{fig:architectures}
\end{figure}

This section describes the details of the network architectures used in our experiments. A detailed description of all the architectures can be found in Fig.~\ref{fig:architectures}
\paragraph{Digits experiments}
For $SVHN \to MNIST$ experiment, we used $DigF1$, $DigC1$, $DigG$ and $DigD$ architectures mentioned in Fig.~\ref{fig:architectures} as our $F$, $C$, $G$ and $D$ networks respectively. For all other digit experiments, we use $DigF2$, $DigC2$, $DigG$ and $DigD$. All models were trained from scratch and were initialized using random Gaussian noise with standard deviation $0.01$. We used Adam solver with base learning rate of $0.0005$ and momentum $0.8$ to train our models. The cost coefficients $\alpha$ and $\beta$ are set as $0.1$ and $0.03$ respectively based on validation splits. We resize all input images to $32 \times 32$ and scale their values to the range $[0,1]$.

\paragraph{OFFICE experiments}
For OFFICE experiments, we used $OfcC$, $OsG$ and $OsD$ architectures mentioned in Fig.~\ref{fig:architectures} as our $C$, $G$ and $D$ networks respectively. The $F$ network is initialized with pretrained Resnet50 model trained on ImageNet, the last layer of which is removed and the resulting $2048$ dimensional vector is used as the feature embedding. We use Adam solver for optimization with a base learning rate of 0.0004 and momentum 0.7 for all the experiments. The dimension of the random noise vector is set as $128$ and the cost coefficient $\alpha$ and $\beta$ are both set as $0.01$.

\paragraph{Synthetic to Real experiments}
Similar to OFFICE experiments, we used $SynC$, $OsG$ and $OsD$ architectures mentioned in Fig.~\ref{fig:architectures} as our $C$, $G$ and $D$ networks respectively. We remove the last layer of the pretrained VGG16 model trained on Imagenet, and initialize it as our $F$ network. The resulting $4096$ dimensional vector is used as the feature embedding. For all the experiments, we used the same hyperparameter settings as those used in the Office experiments.

\section{Noise Analysis}
\begin{figure}[!h]
\centering
\includegraphics[scale=0.5]{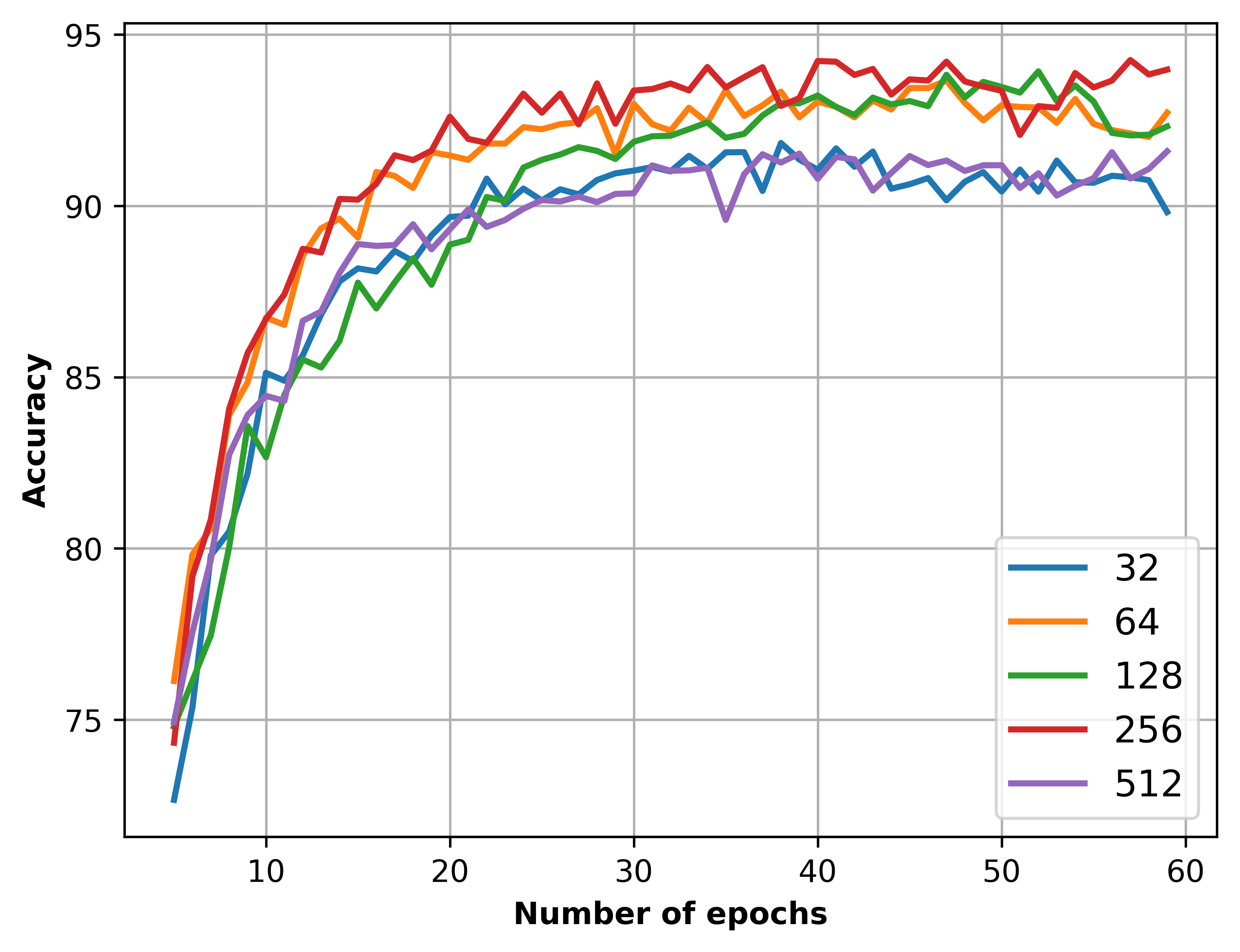}
\caption{Effect of the noise dimension on classification accuracy for the transfer task SVHN $\rightarrow$ MNIST}
\label{fig:noise}
\end{figure}
As described in our approach in the main paper, the input to the generator network $G$ is $x_{g}=[F(x),z,l]$, a concatenated version of the feature embedding, noise vector $z \in \mathbb{R}^{d}$ sampled from $\mathcal{N}(0,1)$ and $l$, the one-hot encoding of the class label. In this section, we perform a study of how the dimensionality of the noise vector $z$ affects the transfer accuracy. In figure \ref{fig:noise}, the transfer accuracy for the task SVHN $\rightarrow$ MNIST is plotted against the number of training epochs. The dimensionality $d$ is varied over the set: $\{32,64,128,256,512\}$. The following observations can be made: (1) The approach is not overly sensitive to $d$, given that all values obtain an average performance of 90.5\% or more. (2) The values of dimensionality that is too low (32) or too high (512) result in slightly suboptimal performance.


\section{Generation visualization}
\begin{figure}[!htb]
\centering
\includegraphics[width=0.37\textwidth]{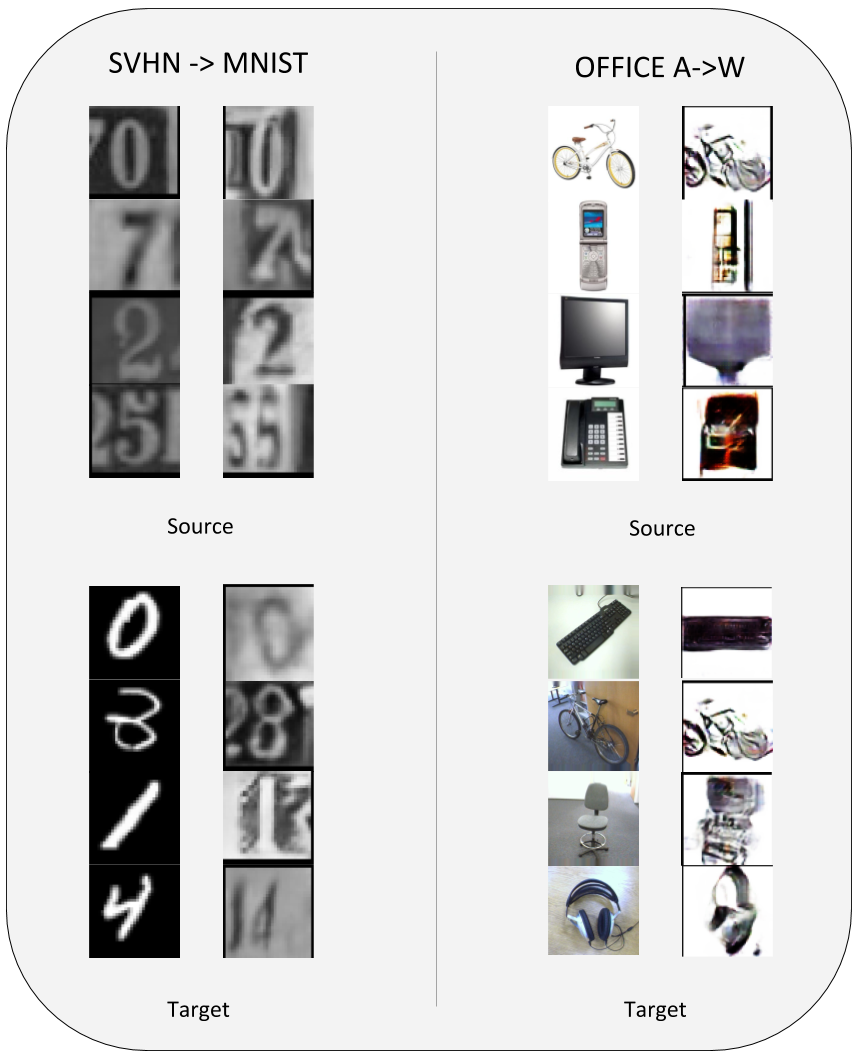}
\caption{Example of images sampled from $G$ after training. In each set, the images on the left indicate the source images and the images on the right indicate the generated images}
\label{fig:generations}
\end{figure}

In Fig.~\ref{fig:generations}, we show some sample images generated by the $G$ network in two experimental settings - $SVHN \to MNIST$ and Office $A \to W$. The top set of images show the generations when the input to the system are the samples taken from the source dataset, while the bottom set are the generations when inputs are the images from the target dataset. We make the following observations: (1) The quality of image generation is better in the digits experiments compared to the Office experiments (2) The generator is able to produce source-like images for both the source and target inputs in a class-consistent manner (3) There is mode collapse in the generations produced in the Office experiments.

The difficulty of GANs in generating realistic images in the Office and Synthetic to real datasets makes it significantly hard for the methods that use
cross-domain image generation as a data augmentation step. Since we rely on the image generation as a mode for deriving rich gradients to the feature extraction network, our method works well even in the presence of severe mode collapse and poor generation quality.  

\begin{table}[!th]
\centering
\label{tab:syn2real}
\caption{Accuracy (mean $\pm$ std\%) values over five independent runs on the Synthetic to real dataset. The best numbers are indicated in \textbf{bold}.}
\label{tab:syn2real_table}
\resizebox{0.45\textwidth}{!}{%
\begin{tabular}{c|c}
\hline
\rule{0pt}{3ex}
Method & CAD $\rightarrow$ PASCAL \\
\hline
\hline 
\rule{0pt}{3ex}
ResNet50 - Source only & 30.2 $\pm$ 0.6 \\
RevGrad & 41.7 $\pm$ 1.3  \\
Ours &  \textbf{46.5} $\pm$ 0.9 \\
\hline
\end{tabular}
}
\end{table}

\section{Synthetic to Real adaptation with ResNet}
This experiment is an extension to the Synthetic to Real experiments in the main paper. Instead of initializing $F$ network with the pretrained VGG16 model, we initialize it with pretrained Resnet-50 model trained on ImageNet as done in the OFFICE experiments. The results of the experiments are presented in Table.~\ref{tab:syn2real_table}. We observe that the model trained only on source domain achieves $30.2\%$ performance, which is $7.9\%$ less than the VGG16 baseline performance mentioned in the main paper. However, our method achieves a performance of $46.5\%$ (which is $16.3\%$ above the baseline) and outperforms other compared approaches.



{
\bibliographystyle{ieee}
\bibliography{egbib}
}
\end{document}